# SENSE-STEP: Learning Sim-to-Real Locomotion for a Sensory-Enabled Soft Quadruped Robot

Storm de Kam[1], Ebrahim Shahabi[1], Cosimo Della Santina[1,2]

*Abstract* — Robust closed-loop locomotion remains challenging for soft quadruped robots due to high dimensions, actuator hysteresis, and difficult-to-model contact dynamics, while conventional proprioception provides limited information about ground interaction. In this paper, we address this challenge with a learning-based control framework for pneumatically actuated soft quadruped equipped with tactile suction-cup feet and validate the approach experimentally on physical hardware. The control policy is trained in simulation through a staged learning process that builds on a reference gait and is progressively refined under randomized environmental conditions. The resulting controller maps proprioceptive and tactile feedback to coordinated pneumatic actuation and suction cup commands, enabling closed-loop locomotion on flat and inclined surfaces. When deployed on the real robot, the closed-loop policy outperforms open-loop baseline, increasing forward speed by 41% on a flat surface and by 91% on a 5° incline. Ablation studies further demonstrate the role of tactile force estimates and inertial feedback in stabilizing locomotion, with performance improvements of up to 56% compared to configurations without sensory feedback.

*Index Terms* — Force and Tactile Sensing, Imitation Learning, Modeling, Control and Learning for Soft Robots, Reinforcement Learning.

## I. INTRODUCTION

ROBOTIC quadrupeds are widely used platform for locomotion in complex environments, offering mobility, agility, and robustness across in a wide range of terrain. Rigid quadrupeds have achieved impressive performance in inspection, agriculture, and autonomous exploration [1]–[3], but typically rely on high-stiffness structures, precise actuation, and carefully tuned control pipelines [4]–[6]. These requirements limit their adaptability and safety in contact-rich or fragile environments.

Soft-legged quadrupeds offer an alternative approach. Their compliant structures enable safe physical interaction, passive impact absorption, and natural adaptation to uneven terrain. These properties make soft quadrupeds promising candidates for applications requiring safe, contact-rich mobility. However, these advantages come with significant control challenges. Soft structures deform in complex, high-dimensional ways that are difficult to model, and their dynamics exhibit strong nonlinearities and hysteresis. Conventional proprioceptive sensors, such as joint encoders, are insufficient for soft robots, and rigid kinematic models cannot accurately represent their deformations. These factors make closed-loop locomotion a persistent open problem.

Prior work has made progress toward this goal. Model-based strategies have been used to stabilize bending and stepping motions in soft legs [7]–[11], but they remain highly sensitive to modeling inaccuracies. Reinforcement learning (RL) approaches have been explored on tendon-driven quadrupeds [12], [13] and in soft-body simulation environments [14], [15], and have enabled locomotion for soft crawling robots [16]. While these results illustrate the potential of learning-based control, they are either confined to simulation or operate on hardware with substantially lower actuation and morphological complexity than full soft quadrupeds.

Despite these advances, experimental demonstrations of learning-based closed-loop locomotion in soft quadrupeds remain limited, particularly for systems that combine pneumatic actuation, high-dimensional deformation, and tactile feedback for active ground interaction.

In this work, we demonstrate simulation-trained closed-loop locomotion on a pneumatically actuated soft quadruped that integrates omnidirectional soft legs with actively anchoring, tactile suction feet. The learned controller coordinates deformation and attachment using only onboard inertial and tactile feedback, without external sensing or motion capture.

## II. SYSTEM OVERVIEW

The TActile Soft Quadruped (TASQ) is a pneumatically actuated soft quadruped robot designed to study closed-loop locomotion under rich tactile feedback. The platform integrates compliant actuation, active ground anchoring, and distributed sensing into a single system, enabling adaptive locomotion in contact-rich environments without reliance on vision. An overview of the robot design is shown in Fig. 1.

Each of the four legs is actuated by a Pneumatic Single-Structure Omnidirectional Actuator (PSSOA) [17]. Each actuator consists of a single elastomeric body with three internal pneumatic chambers, enabling omnidirectional bending and axial extension and contraction through differential pressurization. Compared to multi-bellow or fiber-reinforced designs, this architecture offers a compact form factor, reduced assembly complexity, and smooth, continuous deformation, making it well-suited for quadrupedal locomotion with large workspace requirements.

The PSSOAs' intrinsic compliance provides passive shock absorption and allows the legs to naturally conform to uneven

S. de Kam is with the Cognitive Robotics Department, Delft University of Technology, Mekelweg 2, 2628 CD Delft, the Netherlands. E. Shahabi is with the Cognitive Robotics Department, Delft University of Technology, Mekelweg 2, 2628 CD Delft, the Netherlands (Corresponding author, E.Shahabishalghouni@tudelft.nl). C. Della Santina is with the Cognitive Robotics Department, Delft University of Technology, Mekelweg 2, 2628CD Delft, the Netherlands. C. Della Santina is with the Institute of Robotics and Mechatronics, German Aerospace Center (DLR), 82234 Wesling, Germany ( C.DellaSantina@tudelft.nl).

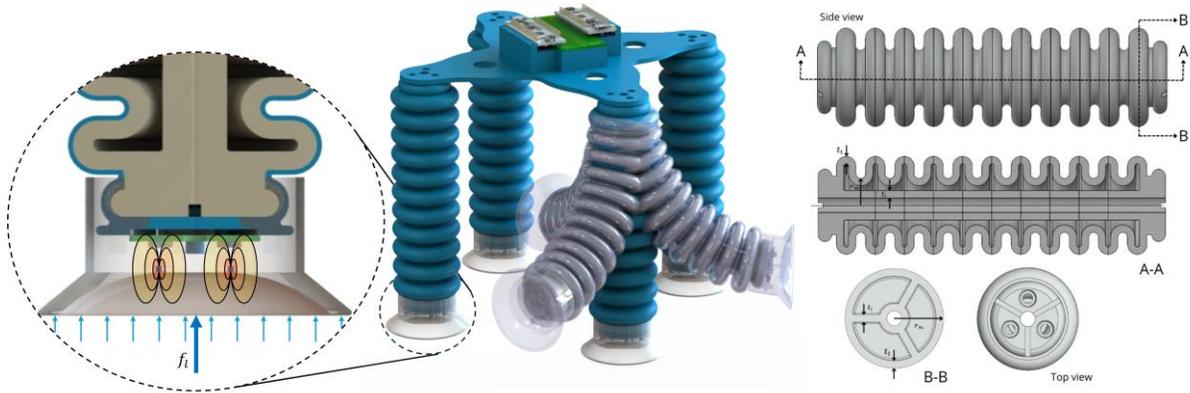

Fig. 1. A general view of the TActile Soft Quadruped (TASQ) robot. Each leg is pneumatically actuated to enable bending and extension for omnidirectional locomotion. A tactile suction cup at the distal end of each leg provides active ground anchoring and estimates ground reaction forces using integrated Hall-effect sensors and embedded magnets.

terrain. All actuators are fabricated using PolyJet additive manufacturing, enabling high geometric fidelity and the repeatable integration of complex internal chamber structures. A rigid central body mechanically couples the four legs and houses pneumatic tubing, electronics, and mounting interfaces, while minimizing interference with leg motion.

At the distal end of each leg, TASQ is equipped with a custom-designed tactile suction cup that serves a dual function, including active attachment to the substrate and local ground reaction force sensing. Suction is generated using a controllable vacuum generator, allowing each foot to selectively anchor to the ground during stance. This capability expands the robot's interaction space beyond friction-based contact alone and enables stable locomotion on inclined surfaces.

The suction cups incorporate a compliant tactile-sensing mechanism based on magnetic-field measurements. Three neodymium magnets are embedded within the silicone suction cup and aligned with three Hall-effect sensors mounted on a rigid PCB. As the suction cup deforms under load, the relative position between magnets and sensors changes, producing a variation in the measurable magnetic field. These signals are converted into estimates of the normal ground reaction force using a learned multilayer perceptron, providing continuous feedback on contact conditions that directly informs gait timing and suction activation during closed-loop locomotion. This sensing approach avoids the stiffness and bulk of conventional load cells and is well-suited for integration into soft robotic systems.

In addition to foot-level tactile sensing, the robot is equipped with an inertial measurement unit (IMU) mounted on the central body, providing estimates of orientation and angular velocity. Together, the IMU and tactile sensors form an observation space that captures both global body dynamics and local contact conditions. Importantly, this sensing configuration enables blind locomotion without cameras or external tracking, relying instead on proprioceptive and tactile feedback that is inherently robust to occlusion and environmental variability.

Actuation is realized through proportional pneumatic valves that regulate chamber pressures in the PSSOAs, along with independent control of the vacuum generators for the suction cups. The resulting action space is high-dimensional, combining continuous pressure commands for soft actuation with continuous control of foot anchoring. This combination allows the robot to coordinate leg deformation, timing, and attachment in a tightly coupled manner. Control commands are generated by an off-board computer and transmitted to the robot at a fixed control frequency.

From a control perspective, TASQ can be seen as a high-dimensional, underactuated, compliant dynamical system with nonlinearities, hysteresis, and contact-dependent behavior. Rather than relying on explicit analytical models of soft-body dynamics, the system is designed to expose informative sensory feedback to learning-based controllers. By integrating soft omnidirectional actuation, active suction-based contact, and tactile force sensing into a unified platform, TASQ provides a physically rich testbed for investigating imitation learning and reinforcement learning methods for closed-loop locomotion in soft quadruped robots.

## III. TACTILE SENSING

Stable tactile sensing is essential for legged locomotion over uneven terrain and on inclined surfaces, and during intermittent contact, where proprioceptive sensing alone is insufficient to reliably estimate ground interaction forces. This challenge is particularly important in soft robots, whose compliant bodies undergo large deformation and exhibit contact dynamics that vary strongly with state and landing conditions. This limitation is particularly pronounced in soft quadrupeds, where large, state-dependent deformations decouple actuation commands from actual foot–ground interaction.

Conventional force and torque sensors are typically rigid and bulky, making them difficult to integrate into soft robotic structures without compromising compliance or increasing system complexity. For this reason, tactile sensing approaches that are compact, deformable, and mechanically compatible

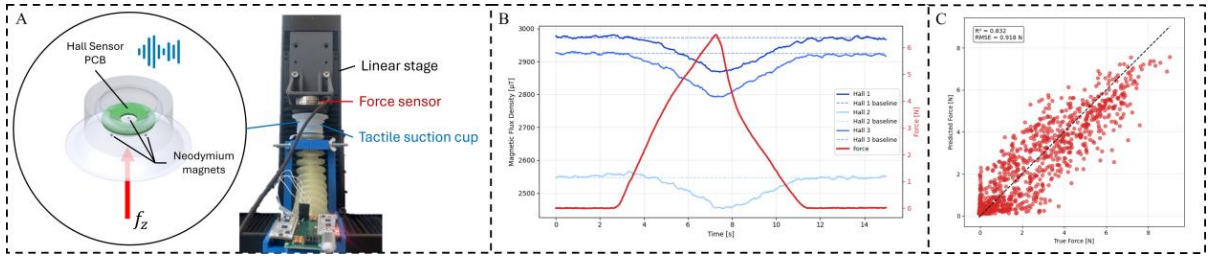

Fig. 2. Overview of the tactile suction cup sensor and calibration procedure. (A) Integrated tactile suction cup design with embedded magnets and Hall-effect sensors, and the experimental calibration setup using a linear stage and reference force sensor. (B) Representative Hall sensor signals and corresponding reference force during a single press cycle. (C) Force estimation performance of the trained MLP on held-out validation data.

with soft materials are more suitable for soft-legged locomotion.

To enable force-aware locomotion in a soft quadruped of realistic complexity, we developed a compact tactile sensor integrated directly into each suction cup, as shown in Fig. 2A. The sensor consists of three Hall-effect sensors mounted on a rigid printed circuit board (PCB), each aligned with a neodymium magnet embedded in the surrounding silicone structure. Deformation of the silicone suction cup under load changes the relative position between the magnets and sensors, resulting in measurable variations in magnetic flux density, as illustrated in Fig. 2B.

The magnetic field measurements are mapped to estimates of the normal ground reaction force through a data-driven calibration procedure. As shown in Fig. 2A, a linear stage equipped with a reference force sensor repeatedly presses the suction cup, generating paired magnetic field and force measurements. A total of 420 press cycles are recorded across multiple pressing velocities and both full and partial contact conditions, to improve robustness across different contact scenarios. Data from all four tactile suction cups are collected to reduce the influence of manufacturing variability.

A subset of the collected data is used to train a Multilayer perceptron (MLP) that estimates the normal ground reaction force from magnetic field readings, while the remaining data are reserved for evaluation. The trained model provides smooth, continuous force estimates suitable for real-time closed-loop control, with force estimation performance reported in Fig. 2C.

Due to the soft materials' inherent compliance, the tactile sensor exhibits hysteresis and may show small drift under sustained loading. These effects are characteristic of soft robotic systems and reflect the realistic mechanical behavior of the sensing structure during long contact.

## IV. LEARNING-BASED CLOSED-LOOP CONTROL FRAMEWORK

To address the challenges of closed-loop locomotion in a high-dimensional soft quadruped, we adopt a staged learning framework that combines imitation learning (IL) and reinforcement learning (RL) as shown in Fig. 3A. This approach enables safe and sample-efficient acquisition of locomotion behaviors, leverages structured motion priors, and gradually refines controllers toward robust, closed-loop performance.

The core of our framework is a feedback policy that maps onboard inertial and tactile observations to high-level actuation commands, allowing the robot to adapt leg deformation and foot anchoring in response to contact conditions, as shown in Fig. 3B. The policy outputs continuous increments for 12 pneumatic chamber pressures and 4 suction cup commands, which are applied at a low command frequency to facilitate stable long-horizon coordination. These high-level commands are executed incrementally within a finer-grained internal loop to ensure smooth, precise actuation without requiring the policy to operate at rigid controller update rates.

### A. Observation and Action Representation

The policy receives a state vector composed of filtered proprioceptive and exteroceptive signals, including body orientation and acceleration from an inertial measurement unit (IMU), tactile force estimates from our suction-cup sensors, internal pressure and suction states, and relative goal information for directed locomotion. To promote rhythmic coordination and coordination across legs, we also include a central pattern generator (CPG) signal in the state. CPGs have been widely used as motion primitives or latent rhythm generators in locomotion control, and recent work shows that integrating CPGs with RL facilitates learning and coordination of rhythmic gaits in high-dimensional systems [18].

The resulting single-frame state has 34 dimensions. To mitigate partial observability arising from sensor noise and delays and to provide short-term temporal context, the policy operates on a frame-stacked observation consisting of the four most recent state vectors concatenated together, yielding a total observation dimension of 136. This short history of observations helps the policy infer latent dynamics and contacts that cannot be captured from a single instantaneous frame, a practice motivated by prior work showing that finite observation histories can mitigate partial observability in RL [19].

The action space consists of 16 continuous control outputs, including 12 pressure increment commands for the pneumatic actuators and 4 gradient commands for the suction actuators. Representing actions as incremental changes at low frequency simplifies learning by stabilizing the range of outputs and abstracting away high-frequency actuation details.

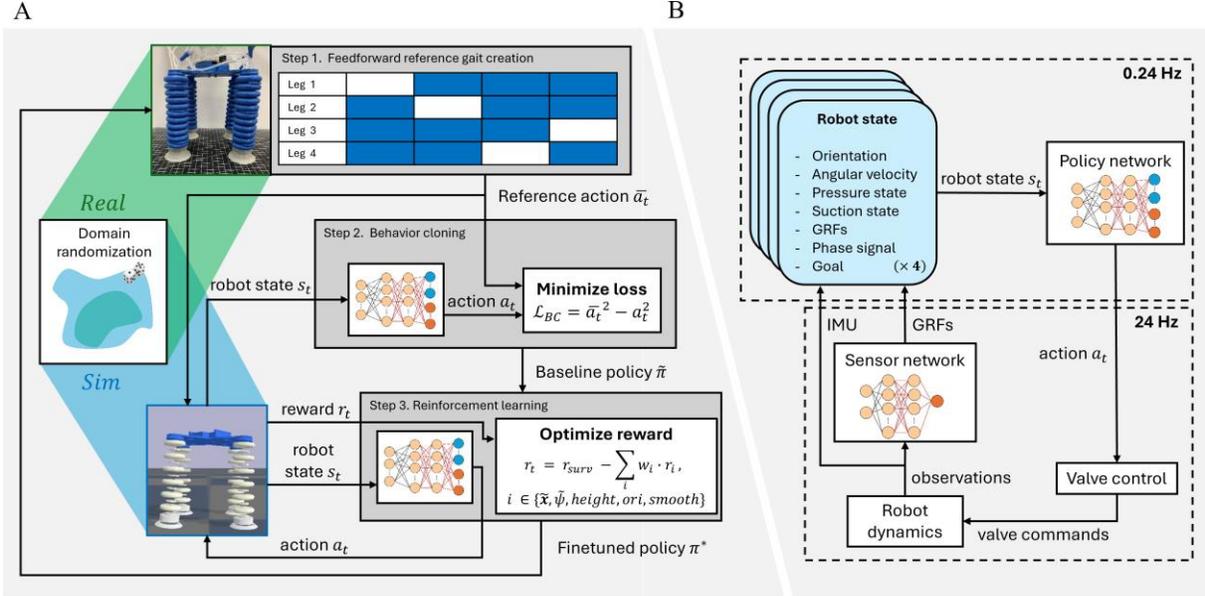

Fig. 3. Overview of the proposed approach. (A) Three-stage policy training pipeline. A heuristic feedforward reference gait is first constructed through iterative experiments on the real robot, inspired by the slow, stable stepping pattern of an elephant. A multilayer perceptron (MLP) is then trained in simulation to imitate this reference gait. The resulting policy is subsequently fine-tuned using reinforcement learning in simulation, with domain randomization applied to improve sim-to-real transfer. (B) Architecture of the learned controller. The policy network outputs high-level actions that are translated into valve commands for actuation. Tactile sensor readings are processed through an MLP to estimate ground reaction forces (GRFs) and fused with IMU measurements to form the robot's state for closed-loop control.

### B. Imitation Learning Warm-Start

Learning locomotion from scratch in high-dimensional soft robots is both sample-inefficient and unstable due to sparse rewards and complex dynamics. To address this, we warm-start RL with behavior cloning (BC) from a reference gait executed on the real robot. Behavior cloning provides an initial policy that captures stable motions and reduces early exploration noise. Combining demonstrations with RL has been shown to significantly accelerate training and improve convergence in robotics tasks, particularly in systems where naive RL struggles with sparse or poorly shaped rewards [20]. The reference gait comprises sequential single-leg steps inspired by slow, stable walking patterns observed in large mammals; this structured prior helps the learned policy begin in a region of the policy space that yields meaningful locomotion behaviors.

### C. Reinforcement Learning for Robust Closed-Loop Locomotion

Once the policy can reliably imitate baseline locomotion in simulation, it is fine-tuned using model-free reinforcement learning to improve robustness, exploit sensory feedback, and achieve goal-directed behaviors. We adopt the Soft Actor–Critic (SAC) algorithm [21] for its demonstrated stability and sample efficiency in continuous, high-dimensional control tasks typical of robotic locomotion and manipulation. SAC's entropy-regularized objective encourages exploration and robustness, which are essential when transferring policies from simulation to hardware.

Initially, the actor's weights are frozen to preserve the learned reference gait and to prevent premature degradation caused by an initially untrained critic. Throughout this warm-up phase, the critic's learning rate gradually decreases. This strategy collectively provides informative samples and enhances early critic learning while protecting the actor from being overridden, thereby improving training stability and convergence.

The reward function $r_t$ balances multiple objectives, including forward progress, stability of body orientation, maintenance of desired body height, and smoothness of commands.

$$r_t = r_{\text{surv}} - \sum_i w_i r_i, \quad i \in \{\tilde{\mathbf{x}}, \tilde{\psi}, \text{h}, \text{ori}, \text{smooth}\} \quad (1)$$

Where $r_{\text{surv}} = 1$ is a constant survival reward applied at each timestep, with episodes terminating when the robot falls. The term $r_{\tilde{\mathbf{x}}} = \mathbf{x}_t - \mathbf{x}_{t-1}$ rewards forward progress along the $x$-direction, while $r_{\tilde{\psi}} = |\psi_t - \psi_{\text{goal}}|^2$ penalizes deviations from the desired yaw orientation. Vertical body stability is encouraged through $r_\text{h} = |z_t - z_{\text{ref}}|$, which penalizes deviations from the target body height, and through $r_{\text{ori}} = |[\phi_t, \theta_t]|^2$, which penalizes pitch and roll deviations. Finally, $r_{\text{smooth}} = |\Delta p_t|^2$ discourages large changes in commanded pressures, promoting smooth actuation.

By combining dense reward terms that encourage forward locomotion, maintain body posture, and regularize actuation, the reward function provides a structured learning signal while still allowing the policy to discover behaviors that surpass the initial reference gait.

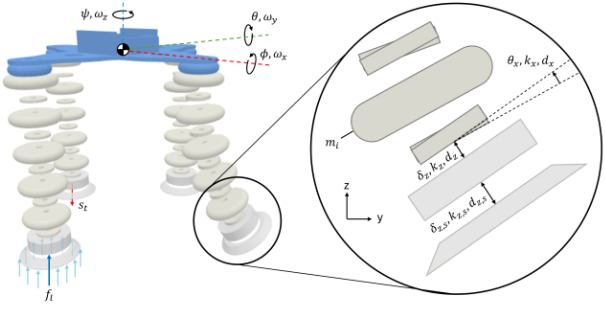

Fig. 4. Robot dynamics model. The dynamics of the legs are approximated with the Pseudo-Rigid-Body Method (PRBM). Revolute and prismatic joints with virtual springs and dampers connect a chain of rigid links.

## V. SIMULATION ENVIRONMENT

### A. Robot Dynamics Model

To simulate the dynamics of the TASQ robot, we use the PyBullet physics engine. As PyBullet provides limited support for deformable bodies, each soft leg is approximated using the *Pseudo-Rigid Body Method (PRBM)* [22], in which the actuator is represented as a chain of rigid links connected by prismatic and revolute joints. This representation captures the dominant bending behavior of the soft actuators while remaining computationally efficient for policy training. To reduce sensitivity to a specific discretization, the number of links per leg is randomized between 5 and 7 during training, thereby improving generalization across actuator variations.

Pneumatic actuation modifies both the equilibrium position and effective stiffness and damping of the actuator. To capture these nonlinear effects, the chamber pressures $\mathbf{p} \in R^K$ are mapped to joint forces $\mathbf{u} \in R^m$ and target joint positions $\mathbf{q}_{\text{targ}} \in R^m$ through:

$$\mathbf{u} = \mathbf{G}\mathbf{p}, \quad \mathbf{q}_{\text{targ}} = \mathbf{K}_p^{-1}\mathbf{u}, \quad (2)$$

where $\mathbf{G} \in R^{m \times K}$ maps chamber pressures to generalized joint forces, and $\mathbf{K}_p \in R^{m \times m}$ is the diagonal matrix of effective joint stiffnesses.

The total stiffness and damping for joint $i$ are:

$$k_{p,\text{tot},i} = k_{p,\text{base}}^{(\tau(i))} + c_1 u_i$$
$$k_{d,\text{tot},i} = k_{d,\text{base}}^{(\tau(i))} + c_2 u_i \quad (3)$$

where $\tau(i)$ maps the joint index $i$ to its type (e.g., revolute $x$-joint), $k_{p,\text{base}}^{(\tau(i))}$ and $k_{d,\text{base}}^{(\tau(i))}$ are nominal base stiffness and damping, and $c_1, c_2$ are scaling coefficients. Per-link gains are obtained via a springs-in-series relation:

$$k_{p,i}^{\text{elem}} = n_i k_{p,\text{tot},i}$$
$$k_{d,i}^{\text{elem}} = n_i k_{d,\text{tot},i} \quad (4)$$

where $n_i$ is the number of links in the $i$-th leg, leading to the diagonal gain matrices $\mathbf{K}_p = \text{diag}(\mathbf{k}_p^{\text{elem}})$ and $\mathbf{K}_d = \text{diag}(\mathbf{k}_d^{\text{elem}})$.

To account for actuator hysteresis, the commanded joint positions $\mathbf{q}^{(t)} \in R^m$ are filtered using a first-order model:

$$\mathbf{q}^{(t)} = \mathbf{q}^{(t-1)} + \boldsymbol{\alpha} \odot \left( \mathbf{q}_{\text{targ}} - \mathbf{q}^{(t-1)} \right), \quad (5)$$

$$\alpha_i = \begin{cases} \alpha_{\text{exp},i} & q_{\text{targ},i} - q_i^{(t-1)} > 0, \\ \alpha_{\text{con},i} & \text{otherwise}, \end{cases} \quad (6)$$

where $\boldsymbol{\alpha} \in R^m$ contains per-joint hysteresis coefficients, with $\alpha_{\text{exp},i}$ and $\alpha_{\text{con},i}$ for expansion/bending and contraction/straightening, respectively.

Suction cups are modeled as two links connected via a prismatic joint. An attractive force toward the ground is applied whenever the vertical distance $d_z$ is below $d_{z,\text{max}}$, the inclination angle $\theta$ is below $\theta_{\text{max}}$, and the suction state $s_{s,i}$ exceeds $s_{\text{min}}$:

$$d_z < d_{z,\text{max}} \ \wedge \ \theta < \theta_{\text{max}} \ \wedge \ s_{s,i} > s_{\text{min}}. \quad (7)$$

A graphical representation of the model is shown in Fig. 4.

### B. Domain Randomization and Observation Noise

To improve sim-to-real transfer and policy robustness, we apply extensive domain randomization during training by perturbing key physical parameters around their nominal values, including link masses $m_i$, joint stiffness $k_j$, joint damping $d_j$, actuator hysteresis $h_\ell$, and suction cup force $s_\ell$. In addition, simulated sensor observations are corrupted with noise to reflect real-world measurement imperfections, affecting the estimated ground reaction force $f_l$, body roll, pitch, and yaw orientations ($\phi$, $\theta$, $\psi$), and angular velocities ($\omega_x$, $\omega_y$, $\omega_z$). These randomized dynamics and noisy observations encourage the policy to generalize beyond the nominal model, supporting stable closed-loop locomotion when deployed on the physical TASQ robot.

## VI. RESULTS

### A. Reinforcement Learning in Simulation

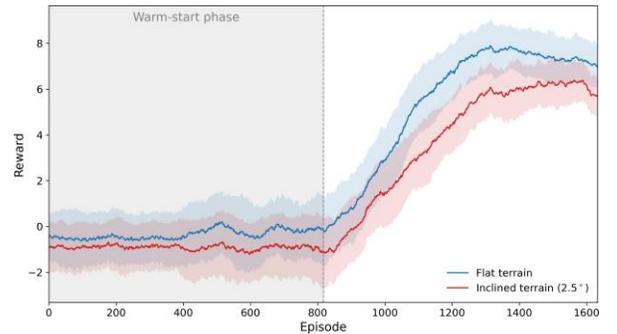

Fig. 5. Average reinforcement learning reward over three random seeds for flat and inclined environments. During the initial warm-up phase, the actor remains frozen while the critic learns. Once the actor is unfrozen, joint learning leads to a steady increase in rewards until convergence.

The policy was first trained in simulation on flat terrain and subsequently refined on a 2.5° incline. As shown in Fig. 5, the episodic reward steadily increases after the warm-start phase and converges after approximately 1300 episodes. Locomotion speed improved by roughly 175% over the behavior-cloned reference gait on flat terrain (Fig. 7). The incline-trained policy achieved an additional 14% improvement on the 2.5° slope and generalized effectively to flat terrain, demonstrating robustness across different inclinations.

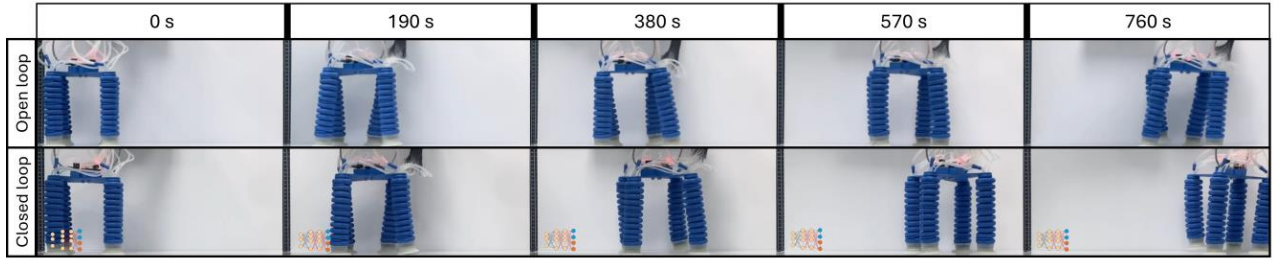

Fig. 6. Comparison of open-loop (top row) and learned closed-loop locomotion (bottom row) on flat terrain. The closed-loop policy covers greater distance, illustrating improved locomotion performance.

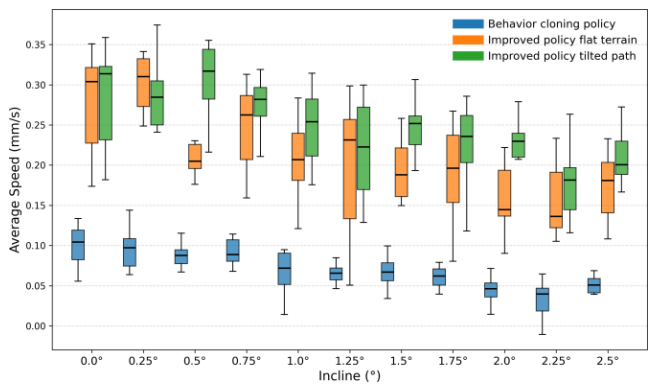

Fig. 7. Simulation results showing locomotion speed across inclinations for the behavior-cloned and reinforcement learning-trained policies. RL-trained policies outperform the baseline consistently.

### B. Experimental Validation on the Physical Robot

The learned policies were deployed on the physical robot. Body trajectories were extracted from video-based motion tracking to quantify performance. On flat terrain, closed-loop control increased forward speed by 41% relative to the open-loop baseline, while also improving body stability and step consistency. Although absolute speeds are limited by pneumatic hysteresis, these results highlight the role of feedback in improving locomotion efficiency. (Fig. 9).

On a 5° incline, the baseline gait speed was 0.116 mm/s, while the learned policy increased speed by 91%. Fig. 8 shows still frames, showing increased locomotion speed as well as body stabilization under the learned policy. The robot actively compensates for the surface inclination by adjusting its body posture to maintain a horizontal orientation.

### C. Contribution of Tactile Feedback and Suction Anchoring

Ablation studies were performed to assess the contribution of sensors and suction cups. Replacing IMU and ground reaction force (GRF) signals with random noise significantly reduced locomotion performance. Enabling closed-loop control increased locomotion speed by 17% on flat terrain and 56% on the incline, demonstrating that tactile and proprioceptive sensing can improve soft-legged locomotion. Disabling suction entirely prevented forward motion on both flat and inclined surfaces. While suction provides the physical anchoring necessary for locomotion, the ablation results indicate that tactile and inertial feedback are critical for coordinating attachment timing and body stabilization, rather than suction alone.

## VII. DISCUSSION

While the proposed learning-based framework successfully achieves closed-loop locomotion on the TASQ robot, several factors limit performance. Crucially, the large hysteresis of the pneumatic actuators slows the hardware response, constraining achievable gait speeds. Testing the framework on faster hardware with reduced hysteresis would allow evaluation of the approach under higher control frequencies and potentially enable more dynamic locomotion.

Future work will extend the framework to uneven terrain, partial foot failure, and external disturbances, which are particularly relevant for soft quadrupeds that rely on active anchoring and tactile feedback for stability. These scenarios will provide controlled benchmarks for evaluating robustness and generalization of the learned controller beyond the flat and inclined surfaces considered here.

## VIII. CONCLUSION

We presented a learning-based control framework for a tactile soft quadruped with novel suction-cup sensors. The approach combines behavior cloning from a reference gait with domain-randomized reinforcement learning, enabling safe exploration and effective policy refinement in simulation.

On the physical robot, closed-loop policies outperformed open-loop control, increasing flat-terrain speed by 41% and incline speed by 91%. Ablation studies confirmed the critical role of IMU and tactile feedback, which improved performance by 17% on flat terrain and 56% on slopes. The controller also stabilized body posture, maintaining near-horizontal orientation during locomotion.

These results demonstrate that integrating tactile feedback with learning-based control enables adaptive, simulation-trained policies to transfer effectively to real soft robots, offering a foundation for future work on more complex terrains and enhanced closed-loop behaviors.

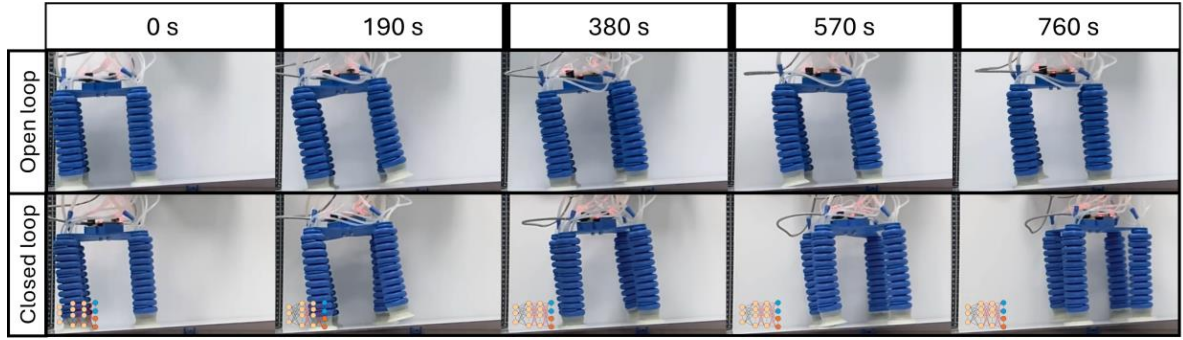

Fig. 8. Comparison of open-loop (top) and learned closed-loop (bottom) locomotion on an inclined surface. The closed-loop policy stabilizes the body and improves forward displacement.

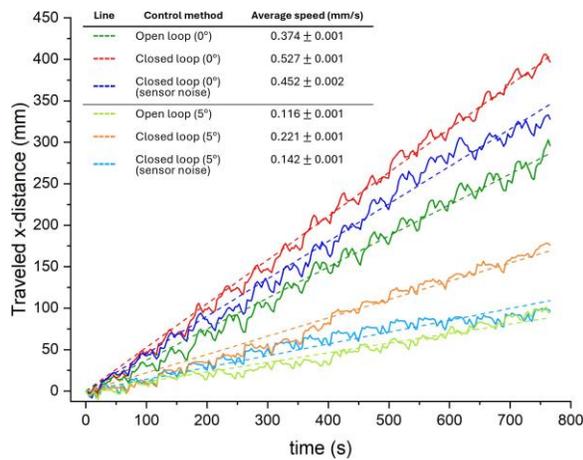

Fig. 9. Traveled distance comparison under open-loop and learned closed-loop control for flat and $5°$ inclined surfaces. Closed-loop control improves forward displacement. Randomizing sensor readings drastically reduces performance, highlighting the importance of feedback.

## REFERENCES

[1] Y. Jang, W. Seol, K. Lee, K.-S. Kim, and S. Kim, "Development of quadruped robot for inspection of underground pipelines in nuclear power plants," *Electronics Letters*, vol. 58, no. 6, pp. 234–236, 2022.

[2] T. Miki, J. Lee, J. Hwangbo, L. Wellhausen, V. Koltun, and M. Hutter, "Learning robust perceptive locomotion for quadrupedal robots in the wild," *Science robotics*, vol. 7, no. 62, p. eabk2822, 2022.

[3] C. Quail, E. Emonot-de Carolis, and F. A. Cheein, "Legged robots in the agricultural context: Analysing their traverse capabilities and performance," in *IECON 2023-49th Annual Conference of the IEEE Industrial Electronics Society*. IEEE, 2023, pp. 01–07.

[4] V. Atanassov, J. Ding, J. Kober, I. Havoutis, and C. Della Santina, "Curriculum-based reinforcement learning for quadrupedal jumping: A reference-free design," *IEEE Robotics & Automation Magazine*, 2024.

[5] Y. Fan, Z. Pei, C. Wang, M. Li, Z. Tang, and Q. Liu, "A review of quadruped robots: Structure, control, and autonomous motion," *Advanced Intelligent Systems*, vol. 6, no. 6, p. 2300783, 2024.

[6] H. Kim, H. Oh, J. Park, Y. Kim, D. Youm, M. Jung, M. Lee, and J. Hwangbo, "High-speed control and navigation for quadrupedal robots on complex and discrete terrain," *Science Robotics*, vol. 10, no. 102, p. eads6192, 2025.

[7] J. M. Bern, P. Banzet, R. Poranne, and S. Coros, "Trajectory optimization for cable-driven soft robot locomotion." in *Robotics: Science and Systems*, vol. 1, no. 3, 2019.

[8] D. Drotman, S. Jadhav, D. Sharp, C. Chan, and M. T. Tolley, "Electronics-free pneumatic circuits for controlling soft-legged robots," *Science Robotics*, vol. 6, no. 51, p. eaay2627, 2021.

[9] Z. Liu and K. Karydis, "Position control and variable-height trajectory tracking of a soft pneumatic legged robot," in *2021 IEEE/RSJ International Conference on Intelligent Robots and Systems (IROS)*. IEEE, 2021, pp. 1708–1709.

[10] J. Kim, E. Im, Y. Lee, and Y. Cha, "Quadrupedal robot with tendon-driven origami legs," *Sensors and Actuators A: Physical*, vol. 378, p. 115769, 2024.

[11] D. D. Arachchige, T. Sheehan, D. M. Perera, S. Mallikarachchi, U. Huzaifa, I. Kanj, and I. S. Godage, "Efficient trotting of soft robotic quadrupeds," *IEEE Transactions on Automation Science and Engineering*, 2025.

[12] X. Niu, K. Tan, D. G. Broo, and L. Feng, "Optimal gait control for a tendon-driven soft quadruped robot by model-based reinforcement learning," in *2025 IEEE International Conference on Robotics and Automation (ICRA)*. IEEE, 2025, pp. 9287–9293.

[13] Q. Ji, S. Fu, K. Tan, S. T. Muralidharan, K. Lagrelius, D. Danelia, G. Andrikopoulos, X. V. Wang, L. Wang, and L. Feng, "Synthesising the optimal gait of a quadruped robot with soft actuators using deep reinforcement learning," *Robotics and Computer-Integrated Manufacturing*, vol. 78, p. 102382, 2022.

[14] P. Schegg, E. Ménager, E. Khairallah, D. Marchal, J. Dequidt, P. Preux, and C. Duriez, "Sofagym: An open platform for reinforcement learning based on soft robot simulations," *Soft Robotics*, vol. 10, no. 2, pp. 410–430, 2023.

[15] G. Tiboni, A. Protopapa, T. Tommasi, and G. Averta, "Domain randomization for robust, affordable and effective closed-loop control of soft robots," in *2023 IEEE/RSJ International Conference on Intelligent Robots and Systems (IROS)*. IEEE, 2023, pp. 612–619.

[16] C. Schaff, A. Sedal, and M. R. Walter, "Soft robots learn to crawl: Jointly optimizing design and control with sim-to-real transfer," *arXiv preprint arXiv:2202.04575*, 2022.

[17] M. S. Xavier, C. D. Tawk, Y. K. Yong, and A. J. Fleming, "3d-printed omnidirectional soft pneumatic actuators: Design, modeling and characterization," *Sensors and Actuators A: Physical*, vol. 332, p. 113199, 2021.

[18] G. Bellegarda and A. Ijspeert, "Cpg-rl: Learning central pattern generators for quadruped locomotion," *IEEE Robotics and Automation Letters*, vol. 7, no. 4, pp. 12 547–12 554, 2022.

[19] G. N. Tasse, M. Riemer, B. Rosman, and T. Klinger, "Finding the framestack: Learning what to remember for non-markovian reinforcement learning," in *Finding the Frame Workshop at RLC 2025*.

[20] A. Nair, B. McGrew, M. Andrychowicz, W. Zaremba, and P. Abbeel, "Overcoming exploration in reinforcement learning with demonstrations," in *2018 IEEE international conference on robotics and automation (ICRA)*. IEEE, 2018, pp. 6292–6299.

[21] A. Raffin, A. Hill, A. Gleave, A. Kanervisto, M. Ernestus, and N. Dormann, "Stable-baselines3: Reliable reinforcement learning implementations," *Journal of machine learning research*, vol. 22, no. 268, pp. 1–8, 2021.

[22] H. Marufkhani, M. A. Khosravi, F. Abdollahi, and M. Zareinejad, "Dynamic modeling of a bio-inspired snake-like soft robot using a parallel robotic architecture," *IEEE Access*, vol. 14, pp. 917–938, 2025.